\documentclass[10pt,twocolumn,letterpaper]{article}

\usepackage[pagenumbers]{cvpr} 

\usepackage{colortbl}
\definecolor{oursbg}{RGB}{232,240,252}

\usepackage{cuted}
\makeatletter
\providecommand\@setmarks{}
\makeatother
\usepackage{tabularx}

\definecolor{cvprblue}{rgb}{0.21,0.49,0.74}
\usepackage[pagebackref,breaklinks,colorlinks,allcolors=cvprblue]{hyperref}

\def\confName{CVPR}
\def\confYear{2027}

\title{Get the Next Moment Right: Short-Term Event Planning for Physically Plausible Video Generation}

\author{%
Hao Liu$^{1,3,*}$\quad Xiang Tang$^{2,3,*}$\quad Ruotong Li$^{3,\dagger}$\quad Xiaopeng Fan$^{2}$\quad Yujiu Yang$^{1,\dagger}$\\
$^{1}$Tsinghua University \quad $^{2}$Harbin Institute of Technology \quad $^{3}$Pengcheng Laboratory\\
{\small $^{*}$Equal contribution \quad $^{\dagger}$Corresponding authors}\\
{\small Project page: \url{https://xdlbw.github.io/STEP/}}
}

\begin{document}
\twocolumn[{%
\maketitle
\begin{minipage}{\textwidth}
  \centering
  \includegraphics[width=\textwidth]{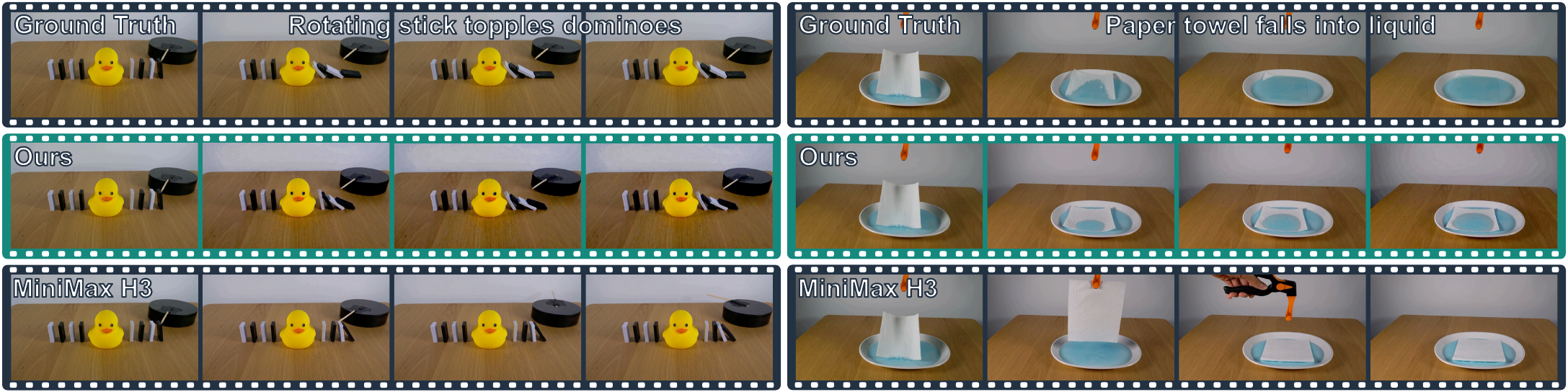}
  \captionof{figure}{Video-to-video (V2V) results on Physics-IQ Verified. Given the same video prefix, rows show the ground truth, ours (STEP on MiniMax H3), and MiniMax H3. Left: a rotating stick topples dominoes. Right: a paper towel falls into liquid.}
  \label{fig:teaser}
\end{minipage}

}]
\begin{abstract}
Video generation models synthesize realistic imagery yet still struggle to produce physically plausible evolution. Existing methods guide generation with a motion trajectory predicted for the entire clip in advance, so prediction errors accumulate and degrade later frames. We propose \textbf{STEP} (\textbf{S}hort-\textbf{T}erm \textbf{E}vent \textbf{P}lanning), a training-free closed-loop framework that guides generation one short clip at a time. Each round consists of four steps: grounding measures the scene state from the latest frames, event planning predicts the next physical event and renders it into a control clip, latent-guided generation produces the next short clip under this control, and verification checks whether the event occurs and resamples the clip if it does not. Because each round is grounded in the frames actually generated, control relies only on reliable short-range predictions. STEP improves the physical plausibility of three frozen generators on Physics-IQ Verified and PhyGenBench, and can be applied to new generators without training.
\end{abstract}
    
\section{Introduction}
\label{sec:intro}

Video generation models now produce videos of high visual quality~\cite{wang2025wan,yang2025cogvideox,tencent2025hunyuanvideo15,teng2025magi1}. They are increasingly used as world simulators~\cite{agarwal2025cosmos,aditi2026cosmos3}, for instance to generate training data~\cite{deng2026rethinking} and to predict action outcomes for embodied agents~\cite{ye2026worldaction}. Such uses require more than realistic appearance: the generated evolution must also be physically plausible, which current models often fail to achieve~\cite{motamed2026physicsiq,meng2025phygenbench,bansal2026videophy2}. For example, a ball should bounce off a wall, a cup should fall when its support is removed, and a paper towel should darken as it absorbs liquid; each outcome depends on the scene conditions~\cite{meng2025phygenbench}. Making generation respect these physical relations remains an important challenge toward reliable world modeling~\cite{kang2025videoworldmodel,esmati2026invisible,chao2026seriality}.

Besides physics-aware post-training~\cite{le2025gravity,kazimi2026hiphy,shen2026phantom,narayanan2026phyco,yuan2026newtongen}, another line of work guides frozen video generators with external physical knowledge. Simulation-guided methods compute physical motion and convert it into generation conditions~\cite{liu2024physgen,tan2026physmotion,foo2026psivg,wang2025physctrl,meng2026phymagic,liu2026realwonder}. Inference-time guidance steers sampling with physical rewards or motion priors~\cite{yuan2026wmreward,han2026physics2steps,tang2026seeking,hassan2026proprio,zhao2026phyrpr}. Agentic methods use language and visual reasoning to construct events, trajectories, or motion programs~\cite{lian2024llmgrounded,xue2025phyt2v,yang2025vlipp,kizil2026lamp,huang2025sketchverify,wang2026eventcausal,feng2026newton,nguyen2026physplan,guan2026agentic}. Although these methods bring physical knowledge into generation, they leave two questions open: \emph{what form the external guidance should take}, and \emph{how far ahead it should be planned}.

\begin{figure}[t]
  \centering
  \includegraphics[width=\linewidth]{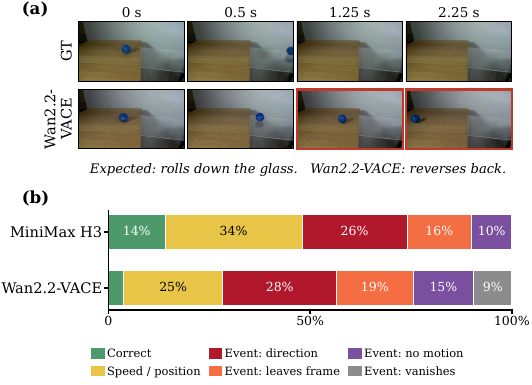}
  \caption{Physical failures of a video generator are mostly wrong or missing events. (a) A ball should roll down the glass panel; Wan2.2-VACE instead reverses it back onto the table (red frames). (b) Failure-type proportions of Wan2.2-VACE and MiniMax H3 on Physics-IQ Verified, computed from 100 manually reviewed samples per generator.}
  \label{fig:failure}
\end{figure}

Answering the first question requires identifying where generated motion goes wrong. We find that most errors originate from wrong events. A generator often misses an expected event or produces a different one, and this wrong event in turn causes errors in speed and position. In Fig.~\ref{fig:failure}(a), a ball should roll down a glass panel, but Wan2.2-VACE~\cite{wang2025wan,jiang2025vace} moves it back onto the table. To measure how common this is, we manually compare 100 videos per generator from Wan2.2-VACE and MiniMax H3~\cite{minimax2026h3} with the ground-truth videos of Physics-IQ Verified~\cite{radsch2026physicsiqverified}; most failures are missing or replaced events (Fig.~\ref{fig:failure}(b)).
External guidance should therefore specify which event each entity should undergo under the current conditions and build motion control around that event.

An intuitive way to apply this is to predict the full motion of each event in advance and use it to control the entire clip. This brings us to the second question: how far ahead control should be planned. However, motion predictions grow less reliable the further ahead they reach, so control based on a full-clip prediction tends to fail in later frames. We therefore keep the requirements of the overall physical process, but plan concrete motion control only for the next short clip, where prediction remains reliable, and update it as generation proceeds.

This idea leads to a loop that alternates planning and generation. Each round plans only the next event and its coarse motion targets. After generation, the new frames are checked for whether the event occurred and re-measured to update the scene state, so the next round starts from actual observations. Events thus play two roles in each round: they set the control target before generation and serve as the check criterion after it.

We introduce \textbf{STEP} (\textbf{S}hort-\textbf{T}erm \textbf{E}vent \textbf{P}lanning), a training-free closed-loop framework for physically plausible video generation with frozen generators (Fig.~\ref{fig:teaser}). STEP first builds a physical specification from the text description and visual context, listing the expected events and the relations they should satisfy. In each round, it plans the next event for a short clip from the latest observations, converts the event targets into a local control signal, and uses this signal to guide generation in latent space. The generated clip is checked against the current event and resampled if the check fails. The retained frames are then re-measured to update the scene state and event progress for the next round. In this way, the specification keeps the whole video consistent with the overall physical process, while each round controls only the next short clip.

Our main contributions are:
\begin{itemize}
    \item \textbf{Event-level control.} We guide generation with short-term physical events instead of full motion trajectories, which directly addresses the missing or replaced events behind most physical failures.
    \item \textbf{Short-term closed-loop planning.} We plan one short clip at a time from the latest observations and resample any clip that fails its event check, which avoids long-range prediction and limits error propagation.
    \item \textbf{Training-free improvement.} As a training-free framework, STEP improves the physical plausibility of different frozen video generators across generation settings.
\end{itemize}

\section{Related Work}
\label{sec:related}

\subsection{Simulation-Guided Video Generation}

Simulation-guided methods compute object motion with a physical solver and convert it into conditions for video generation. PhysGen~\cite{liu2024physgen} simulates rigid-body dynamics in image space, and PhysMotion~\cite{tan2026physmotion} combines a 3D representation with continuum mechanics. PSIVG~\cite{foo2026psivg} reconstructs a scene from an initial video and guides subsequent synthesis with simulated trajectories. Other methods couple simulation with vision-language model (VLM) feedback~\cite{meng2026phymagic}, real-time action control~\cite{liu2026realwonder}, or executable code worlds~\cite{wang2026codeasworlds}. These methods give motion an explicit physical basis, but require scene reconstruction, parameter estimation, and a solver configured for the materials and interactions involved~\cite{wang2025physctrl}. STEP instead plans events with a VLM and requires no physical simulator.

\subsection{Inference-Time Physics Guidance}

Another line of work improves physical plausibility at inference time. WMReward~\cite{yuan2026wmreward} uses a latent world model as a physical reward for candidate search and sampling guidance. Seeking Physics in Diffusion Noise~\cite{tang2026seeking} derives verification signals from intermediate diffusion features for candidate selection and gradient guidance. Physics in 2-Steps~\cite{han2026physics2steps} extracts a motion prior from few-step generation. It then constrains latent frame differences during later denoising to preserve this motion while refining visual details. Other methods refine samples with latent self-scoring~\cite{hassan2026proprio} or separately reasoned physical constraints~\cite{zhao2026phyrpr}. Most of these methods rely on learned rewards or the generator's own motion prior: rewards may mix physical and perceptual factors, and errors in the initial motion prior may persist through refinement. STEP instead verifies each clip against its expected event using states measured from the generated frames.

\begin{figure*}[t]
  \centering
  \includegraphics[width=\textwidth]{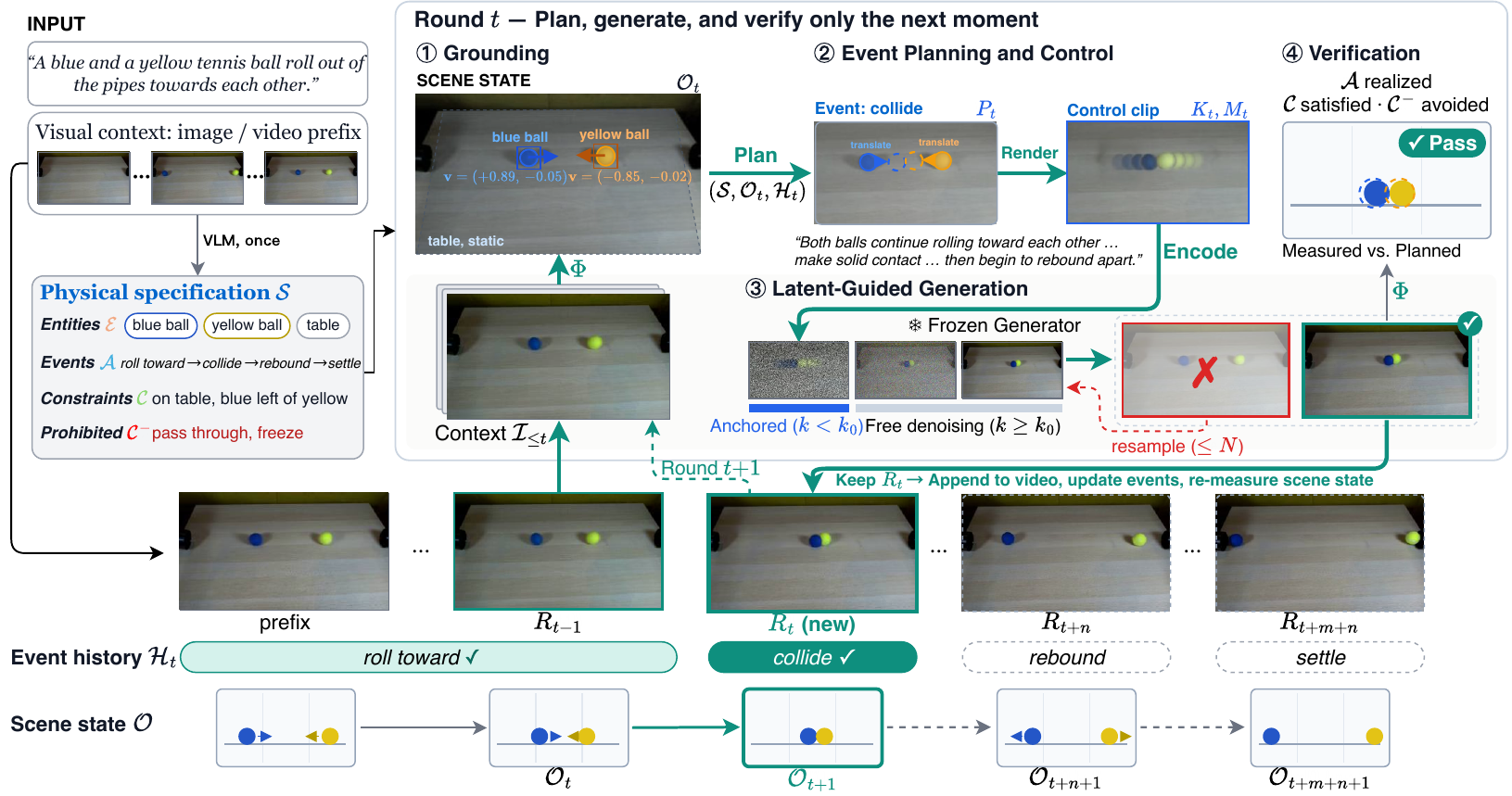}
  \caption{Overview of STEP. Each round (1) grounds the latest frames into a scene state, (2) plans the next event and renders it into a control clip, (3) generates the next short clip with latent-guided generation, and (4) verifies the result. Failed clips are resampled; accepted clips extend the video and update the event history and scene state for the next round; $R_{t+n}$ and $R_{t+m+n}$ ($m,n\geq 1$) denote clips accepted in later rounds. Frames come from a real case; plan targets, the control clip, and state sketches are schematic.}
  \label{fig:method}
\end{figure*}

\subsection{Agentic Planning and Verification}

Agentic methods use language or visual reasoning to turn physical knowledge into explicit plans before generation. These plans differ in granularity. LVD~\cite{lian2024llmgrounded} specifies object boxes frame by frame, VLIPP~\cite{yang2025vlipp} provides a coarse physical motion plan, and LAMP~\cite{kizil2026lamp} compiles language-generated motion programs into explicit trajectories. At a higher level, Chain of Event-Centric Causal Thought~\cite{wang2026eventcausal} reasons over causal event chains, and PhysPlan~\cite{nguyen2026physplan} represents physical changes on specific objects to select the properties and regions to guide. SketchVerify~\cite{huang2025sketchverify} and PhysAgent~\cite{li2026physagent} further verify motion sketches or simulated physics before final synthesis. In most of these methods, the plan is fixed for the entire clip before generation, so control of later frames relies on long-range predictions whose errors carry into the generated video.

Other methods verify generated videos and use the results to refine generation. PhyT2V~\cite{xue2025phyt2v} and Bootstrapping~\cite{liu2025bootstrapping} revise prompts based on generated results, and PhysRepair~\cite{guan2026agentic} diagnoses invalid states and locally repairs the affected content. NEWTON~\cite{feng2026newton} trains a planner that calls physics-aware tools, such as keyframe generation and scientific computation, to build conditioning, and re-plans when a verifier rejects the result. State World Models~\cite{ki2026stateworld} maintains an evolving world state updated with generated clips, and Code World Model~\cite{chen2026codeworldmodel} maintains such a state with a coding agent. Unlike these methods, STEP plans, controls, and verifies each event within the same short clip, and re-plans every clip from the frames just generated.

\section{Method}
\label{sec:method}

\subsection{Overview}

We present \textbf{STEP}, a training-free closed-loop framework for physically plausible video generation conditioned on text, an image, or a video prefix.

STEP first builds a physical specification of the expected events and the relations they should satisfy, and measures the scene state from the latest frames (Sec.~\ref{sec:method_spec}). In each round, it plans the next event for a short clip and converts it into a local control signal (Sec.~\ref{sec:method_plan}). This signal guides generation in latent space, and the generated clip is verified against the event and resampled on failure (Sec.~\ref{sec:method_sampling}). The retained frames are then re-measured to update the scene state for the next round. Figure~\ref{fig:method} illustrates this loop.

\subsection{Physical Specification and Visual Grounding}
\label{sec:method_spec}

When the scene calls for a change, a generated clip must realize it, and the change must be physically consistent with the scene. These two requirements rest on different evidence, so we keep two coupled objects. The physical specification $\mathcal{S}$ states which entities are present, what should happen to them, and which relations must hold while it happens. The scene state $\mathcal{O}_t$ records what the latest frames show. $\mathcal{S}$ contains both expected events and prohibited outcomes. Expected events exclude a clip that matches the prompt but stays static, and prohibited outcomes exclude changes that contradict the material and contact relations of the scene. Both are stated as observable consequences and relations rather than a prescribed trajectory, since fixing a long trajectory in advance would reintroduce the long-range prediction that the closed loop avoids.

$\mathcal{S}$ is built once at initialization by prompting a VLM with the text description and the visual context available at that point: the conditioning image in image-to-video (I2V) generation and the sampled prefix frames in video-to-video (V2V) generation. Text-to-video (T2V) generation has no context of its own; we generate a first frame with GPT-Image-2.5 Flare and thereafter treat the setting as I2V, so grounding always operates on an actual image. The output is
\begin{equation}
\mathcal{S} = (\mathcal{E}, \mathcal{A}, \mathcal{C}, \mathcal{C}^{-}),
\label{eq:physical_specification}
\end{equation}
where $\mathcal{E}$ is the set of relevant entities, each with a name and a short segmentation prompt, $\mathcal{A}$ the expected events, $\mathcal{C}$ the observable constraints, and $\mathcal{C}^{-}$ the prohibited outcomes.

Grounding ties $\mathcal{S}$ to pixels. SAM 3~\cite{carion2026sam3} segments each entity from its name and prompt, and the masks yield the scene state
\begin{equation}
\mathcal{O}_t = \Phi(\mathcal{E}, \mathcal{I}_{\le t}),
\label{eq:observation_state}
\end{equation}
where $\mathcal{I}_{\le t}$ is the latest context window, drawn from the conditioning frames and previously generated frames, and $\Phi$ is the segmentation-and-measurement procedure. $\Phi$ measures position, velocity, visibility, and support relations on the image plane, with velocity taken over consecutive frames. $\mathcal{S}$ remains fixed throughout the video, assuming that the entities and expected events named at initialization stay valid. In contrast, $\mathcal{I}_{\le t}$ is updated and $\Phi$ is re-run every round. The resulting $\mathcal{O}_t$ is the reference for the local checks in Sec.~\ref{sec:method_sampling}, which cover only the operation selected for that round rather than the whole specification.

\subsection{Short-Term Event Planning and Control}
\label{sec:method_plan}

We organize physical intervention around observable events. Each round plans only the next short clip and states an event with a coarse outcome: which entity changes, by which operation, and roughly where that operation ends. Planning receives the specification $\mathcal{S}$, the scene state $\mathcal{O}_t$ measured from the latest context $\mathcal{I}_{\le t}$, that context itself, and the events accepted in earlier rounds, $\mathcal{H}_t$, with $\mathcal{H}_0=\emptyset$:
\begin{equation}
P_t=\operatorname{Plan}(\mathcal{S},\mathcal{O}_t,\mathcal{I}_{\le t},\mathcal{H}_t).
\label{eq:event_plan}
\end{equation}
Here $t$ indexes rounds. $\mathcal{O}_t$ records where things are now, and $\mathcal{H}_t$ records which events have passed the local event checks. Of $\mathcal{I}_{\le t}$, only the most recent frames are read. $P_t$ names the event, the entities it concerns, an operation and a coarse target for each, one sentence of text, and whether region operations can express the change. Operations come from a small vocabulary---none, translate, scale, rotate, appear, disappear, and color---and a coarse target is a box, a scale, a rotation, or a color.

A coarse target names where the entity ends, not how it gets there, yet the control clip needs the entity's position in every frame of the clip. These positions are filled from the measured state. A translation begins at the measured position and continues at the speed the entity already has, so motion carries across the round boundary without a jump. Where a gravity estimate exists, gravity bends the path over the interval. A support restricts the path only where the entity and the support overlap horizontally, and the path follows the support if the support itself moves. In the first round with only a conditioning image, the initial velocity comes from the plan rather than a measurement.

Control construction places the segmented entity regions along these per-frame positions over a background plate, from which the source location and the swept regions are cleared first. It yields the control clip $K_t$ with a region mask $M_t$ that marks where it applies:
\begin{equation}
(K_t,M_t)=\operatorname{Render}(P_t,\mathcal{O}_t,\mathcal{I}_{\le t}).
\label{eq:control_render}
\end{equation}
Static regions are carried over from the last context frame, while unplanned moving regions are left open to generation. When the clip carries an event, or a change no region operation can express, the plan's text enters the prompt.

\begin{table*}[t]
  \centering
  \caption{Results on Physics-IQ Verified for image-to-video (I2V) and video-to-video (V2V) generation. Score: Physics-IQ Verified score; SIoU, STIoU, and WSIoU: spatial, spatiotemporal, and weighted spatial IoU. The symbol ``---'' denotes that V2V is not supported. Bold: best Score within each base-generator group.}
  \label{tab:main_piq}
  \small
  \setlength{\tabcolsep}{4pt}
  \begin{tabular}{l ccccc ccccc}
    \toprule
    & \multicolumn{5}{c}{I2V} & \multicolumn{5}{c}{V2V} \\
    \cmidrule(lr){2-6} \cmidrule(lr){7-11}
    Method & SIoU$\uparrow$ & STIoU$\uparrow$ & WSIoU$\uparrow$ & MSE$\downarrow$ & Score$\uparrow$ & SIoU$\uparrow$ & STIoU$\uparrow$ & WSIoU$\uparrow$ & MSE$\downarrow$ & Score$\uparrow$ \\
    \midrule
    HunyuanVideo 1.5~\cite{tencent2025hunyuanvideo15} & 0.2890 & 0.1502 & 0.1620 & 0.01044 & 31.17 & --- & --- & --- & --- & --- \\
    CogVideoX-5B~\cite{yang2025cogvideox} & 0.1854 & 0.1958 & 0.0901 & 0.00888 & 29.31 & --- & --- & --- & --- & --- \\
    LTX-2.5~\cite{hacohen2026ltx2} & 0.2270 & 0.1508 & 0.1220 & 0.01696 & 25.38 & 0.3915 & 0.2638 & 0.2560 & 0.00701 & 44.67 \\
    LTX-Video 0.9.8 13B~\cite{hacohen2025ltxvideo} & 0.2200 & 0.2034 & 0.1245 & 0.01461 & 26.95 & 0.3399 & 0.2841 & 0.2160 & 0.00915 & 39.89 \\
    \midrule
    Wan2.2-VACE-Fun-A14B~\cite{wang2025wan,jiang2025vace} & 0.2213 & 0.1482 & 0.0968 & 0.01977 & 23.80 & 0.3121 & 0.1917 & 0.1692 & 0.00777 & 35.65 \\
    \quad + WMReward~\cite{yuan2026wmreward} & 0.2907 & 0.1861 & 0.1727 & 0.00937 & 30.62 & 0.3507 & 0.1640 & 0.2049 & 0.00813 & 36.87 \\
    \rowcolor{oursbg} \quad + STEP (ours) & 0.2331 & 0.2403 & 0.1294 & 0.00868 & \textbf{32.36} & 0.3100 & 0.2513 & 0.1830 & 0.00733 & \textbf{38.47} \\
    \midrule
    MAGI-1 4.5B~\cite{teng2025magi1} & 0.1720 & 0.2692 & 0.1022 & 0.01081 & 27.63 & 0.2787 & 0.3112 & 0.1794 & 0.00747 & 37.56 \\
    \quad + WMReward~\cite{yuan2026wmreward} & 0.1440 & 0.3271 & 0.0684 & 0.00783 & 29.20 & 0.2409 & 0.3171 & 0.1432 & 0.00594 & 41.00 \\
    \rowcolor{oursbg} \quad + STEP (ours) & 0.2166 & 0.3206 & 0.1371 & 0.00820 & \textbf{34.34} & 0.3113 & 0.3633 & 0.2168 & 0.00642 & \textbf{43.05} \\
    \midrule
    MiniMax H3~\cite{minimax2026h3} & 0.3047 & 0.1766 & 0.1698 & 0.00662 & 36.42 & 0.3467 & 0.2264 & 0.2135 & 0.00685 & 42.88 \\
    \quad + WMReward~\cite{yuan2026wmreward} & 0.3191 & 0.1848 & 0.1796 & 0.00769 & 36.63 & 0.3745 & 0.2933 & 0.2491 & 0.00516 & 44.47 \\
    \rowcolor{oursbg} \quad + STEP (ours) & 0.3386 & 0.2086 & 0.2047 & 0.00607 & \textbf{38.41} & 0.4172 & 0.2859 & 0.2646 & 0.00541 & \textbf{47.40} \\
    \bottomrule
  \end{tabular}
\end{table*}

\subsection{Latent-Guided Generation and Verification}
\label{sec:method_sampling}

The control clip $K_t$ states how the current event should unfold. Encoding it as $z_t=\operatorname{Enc}(K_t)$, we anchor its regions in the early denoising steps, so the expected change enters the sampling process itself. Let $\tilde{x}_t^{(k+1)}$ be the sample produced by the generator's $(k{+}1)$-th denoising update from $x_t^{(k)}$. For $k<k_0$, its masked part is replaced by the control latent noised to the corresponding level:
\begin{equation}
\begin{aligned}
x_t^{(k+1)}={}&(1-\mathbf{m}_t)\odot\tilde{x}_t^{(k+1)}\\
&+\mathbf{m}_t\odot\big[(1-\sigma_{k+1})z_t+\sigma_{k+1}\epsilon_t\big].
\end{aligned}
\label{eq:latent_anchor}
\end{equation}
Here $\odot$ denotes element-wise multiplication, $\mathbf{m}_t$ is $M_t$ mapped to latent resolution, with the conditioning frames excluded, $\sigma_{k+1}$ the noise level after the update, which decreases from $1$ (pure noise) toward $0$ (clean latent) as $k$ grows, and $\epsilon_t$ the initial noise shared by all denoising steps in round $t$, so that $x_t^{(0)}=\epsilon_t$. Early steps set the overall layout while detail is still forming, so an anchor placed there organizes the surrounding content rather than overwriting it. The mask covers the planned regions together with the static background, which the control clip inherits from the last context frame. Anchoring both keeps the anchored content consistent with the background it continues, and the rest is left to the generator. Anchoring stops after $k_0$ updates, after which the sample is no longer replaced and denoising proceeds on its own.

After generation, the operation of this round decides what to check: whether the planned change occurred, and whether the constraints in $\mathcal{C}$ and $\mathcal{C}^{-}$ that this operation affects hold. Checks are measured by $\Phi$ on the generated clip; a translation is checked against the per-frame positions used to build $K_t$, rather than the planner's endpoint. A check returns pass, fail, or unmeasurable; unmeasurable is not counted as pass. On failure, the clip is resampled with a new seed for up to $N$ additional attempts, keeping the plan and context fixed, and if all attempts fail, the candidate whose measured outcome is closest to the plan is accepted.

The accepted clip $R_t$ closes the loop:
\begin{equation}
\begin{aligned}
\mathcal{I}_{\le t+1}&=\operatorname{Update}(\mathcal{I}_{\le t},R_t),\\
\mathcal{H}_{t+1}&=\mathcal{H}_t\cup\mathcal{A}_t^{\mathrm{acc}},\\
\mathcal{O}_{t+1}&=\Phi(\mathcal{E},\mathcal{I}_{\le t+1}),
\end{aligned}
\label{eq:feedback_update}
\end{equation}
where $\operatorname{Update}$ appends the accepted frames and retains the latest context window, and $\mathcal{A}_t^{\mathrm{acc}}$ is the set of events whose checks pass for all planned entities in the accepted candidate; events that do not pass remain available to later rounds. Each round commits to one short moment, and the next round starts from what that moment shows, so short interventions accumulate into a full video.

\section{Experiments}
\label{sec:experiments}

\begin{figure*}[t]
  \centering
  \includegraphics[width=\textwidth]{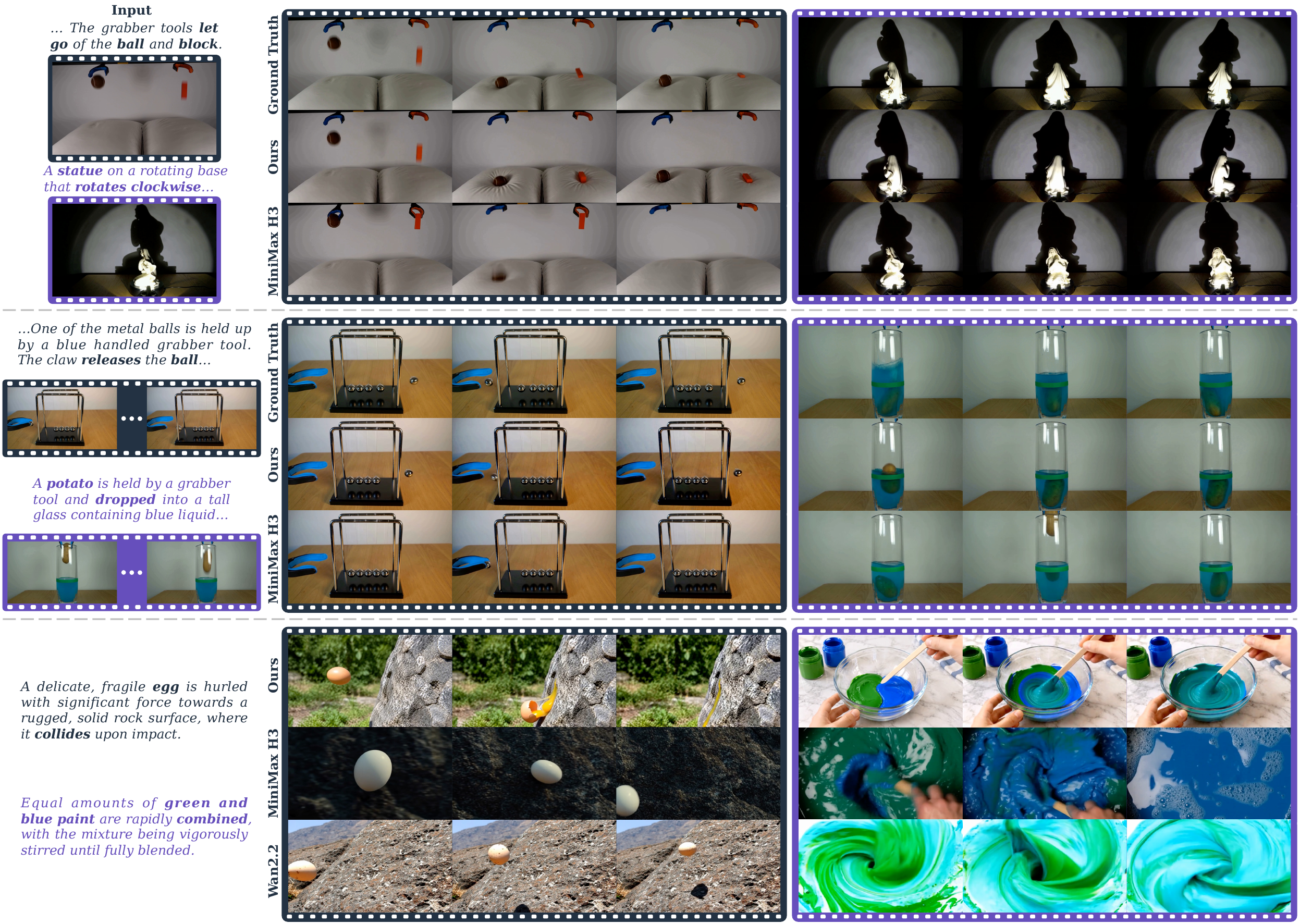}
  \caption{Qualitative comparison. Rows from top to bottom show image-to-video (I2V), video-to-video (V2V), and text-to-video (T2V) generation, with two cases per row and three generated frames per method. I2V and V2V compare the ground truth, ours (STEP on MiniMax H3), and MiniMax H3 given the same visual input. T2V compares ours (STEP on MiniMax H3), MiniMax H3, and Wan2.2-VACE given the same prompt.}
  \label{fig:qualitative}
\end{figure*}

\subsection{Experimental Setup}
\label{sec:exp_setup}

\noindent\textbf{Tasks and benchmarks.}
We evaluate three tasks. In I2V generation, the model receives a text description and a conditioning image; in V2V generation, it receives the text description and a video prefix. We evaluate both on Physics-IQ Verified~\cite{radsch2026physicsiqverified}, which contains 66 scenes recorded from three viewpoints, giving 198 samples. In V2V, the prefix lasts 3 seconds and the generated continuation 5 seconds. For T2V generation, we use PhyGenBench~\cite{meng2025phygenbench}, which contains 160 prompts covering 27 physical laws. STEP first generates an initial image from each prompt with GPT-Image-2.5 Flare and then generates the video from this image as in I2V.

\noindent\textbf{Baselines.}
We compare STEP with two groups of methods. For each of the three base generators, we report the generator alone and with WMReward~\cite{yuan2026wmreward}, an inference-time physics guidance method; WMReward follows the settings of its original paper and released code. We also report other open-source video generators as reference points: HunyuanVideo 1.5~\cite{tencent2025hunyuanvideo15}, CogVideoX-5B~\cite{yang2025cogvideox}, LTX-2.5~\cite{hacohen2026ltx2}, and LTX-Video 0.9.8~\cite{hacohen2025ltxvideo}, as well as NewtonGen~\cite{yuan2026newtongen} on PhyGenBench.

\noindent\textbf{Evaluation metrics.}
On Physics-IQ Verified, we report the Physics-IQ Verified score and its component metrics, which are based on intersection over union (IoU) and mean squared error (MSE): spatial IoU (SIoU), spatiotemporal IoU (STIoU), weighted spatial IoU (WSIoU), and MSE. On PhyGenBench, we follow the closed branch of the official PhyGenEval protocol but replace its GPT-4o judge with GPT-6 Astra, and evaluate all methods in Tab.~\ref{tab:main_phygen} with the same judge. We report the overall Physical Commonsense Alignment (PCA) score and the scores of its three evaluation stages: key phenomena detection, physics order verification, and overall naturalness. We also report the PCA score in each of the four physical domains: mechanics, optics, thermal, and material. All PhyGenBench scores are normalized to $[0,1]$.

\noindent\textbf{Implementation details.}
We apply STEP to three base generators: Wan2.2-VACE-Fun-A14B~\cite{wang2025wan,jiang2025vace}, MAGI-1 4.5B~\cite{teng2025magi1}, and MiniMax H3~\cite{minimax2026h3}. The VLM in STEP is GPT-6 Astra with medium reasoning effort for all generators and tasks. Every generator stays frozen and runs through its released inference implementation with the default or recommended settings; since these interfaces differ, we do not impose shared sampler parameters. The clip length, the number of anchored denoising steps $k_0$, the maximum number of resampling attempts $N$, and other task- and model-specific configurations are listed in the supplementary material.

\begin{table*}[t]
  \centering
  \caption{Results on PhyGenBench for text-to-video (T2V) generation, with all scores normalized to $[0,1]$. Stage: Key Phen., Order, and Natural. denote key-phenomena detection, physics-order verification, and overall naturalness; PCA is the overall Physical Commonsense Alignment score. Domain: PCA score within each physical domain. Bold: best value in each column within each base-generator group.}
  \label{tab:main_phygen}
  \small
  \setlength{\tabcolsep}{4pt}
  \begin{tabular}{l cccc cccc}
    \toprule
    & \multicolumn{4}{c}{Stage} & \multicolumn{4}{c}{Domain} \\
    \cmidrule(lr){2-5} \cmidrule(lr){6-9}
    Method & Key Phen.$\uparrow$ & Order$\uparrow$ & Natural.$\uparrow$ & PCA$\uparrow$ & Mechanics$\uparrow$ & Optics$\uparrow$ & Thermal$\uparrow$ & Material$\uparrow$ \\
    \midrule
    HunyuanVideo 1.5~\cite{tencent2025hunyuanvideo15} & 0.6563 & 0.4167 & 0.5063 & 0.5146 & 0.5333 & 0.5867 & 0.4444 & 0.4583 \\
    CogVideoX-5B~\cite{yang2025cogvideox} & 0.6771 & 0.4479 & 0.4563 & 0.5250 & 0.4417 & 0.6467 & 0.5000 & 0.4750 \\
    LTX-2.5~\cite{hacohen2026ltx2} & 0.6333 & 0.4146 & 0.4833 & 0.5083 & 0.5417 & 0.6067 & 0.3667 & 0.4583 \\
    LTX-Video 0.9.8 13B~\cite{hacohen2025ltxvideo} & 0.5708 & 0.3271 & 0.4313 & 0.4500 & 0.4417 & 0.5400 & 0.4000 & 0.3833 \\
    NewtonGen~\cite{yuan2026newtongen} & 0.6354 & 0.4125 & 0.4354 & 0.5104 & 0.4667 & 0.5933 & 0.5222 & 0.4417 \\
    \midrule
    Wan2.2-VACE-Fun-A14B~\cite{wang2025wan,jiang2025vace} & 0.6250 & 0.4188 & 0.4729 & 0.4958 & 0.4583 & 0.6267 & 0.3667 & 0.4667 \\
    \quad + WMReward~\cite{yuan2026wmreward} & 0.6188 & 0.3938 & 0.5188 & 0.4667 & 0.4667 & 0.6000 & 0.2556 & 0.4583 \\
    \rowcolor{oursbg} \quad + STEP (ours) & \textbf{0.7363} & \textbf{0.4451} & \textbf{0.7152} & \textbf{0.6354} & \textbf{0.6583} & \textbf{0.7619} & \textbf{0.4889} & \textbf{0.5641} \\
    \midrule
    MAGI-1 4.5B~\cite{teng2025magi1} & 0.6458 & 0.3896 & 0.4792 & 0.5083 & 0.4750 & 0.6467 & \textbf{0.4222} & \textbf{0.4333} \\
    \quad + WMReward~\cite{yuan2026wmreward} & 0.6479 & 0.3146 & 0.4083 & 0.4958 & 0.4417 & 0.6400 & \textbf{0.4222} & 0.4250 \\
    \rowcolor{oursbg} \quad + STEP (ours) & \textbf{0.6667} & \textbf{0.4479} & \textbf{0.4896} & \textbf{0.5479} & \textbf{0.4917} & \textbf{0.7733} & 0.4000 & \textbf{0.4333} \\
    \midrule
    MiniMax H3~\cite{minimax2026h3} & 0.6833 & 0.4917 & 0.5438 & 0.5792 & 0.5667 & 0.6933 & 0.5444 & 0.4750 \\
    \quad + WMReward~\cite{yuan2026wmreward} & 0.6667 & 0.5229 & 0.5375 & 0.5854 & 0.5500 & 0.7067 & 0.5556 & 0.4917 \\
    \rowcolor{oursbg} \quad + STEP (ours) & \textbf{0.7000} & \textbf{0.6146} & \textbf{0.6083} & \textbf{0.6417} & \textbf{0.6000} & \textbf{0.7600} & \textbf{0.6444} & \textbf{0.5333} \\
    \bottomrule
  \end{tabular}
\end{table*}

\subsection{Results}
\label{sec:exp_main}

\subsubsection{Image- and Video-Conditioned Generation}
\label{sec:exp_visual}

\noindent\textbf{Quantitative results.}
Table~\ref{tab:main_piq} reports results on Physics-IQ Verified. STEP improves the score of all three base generators in both settings. The base generators often miss an expected event or produce a different one (Fig.~\ref{fig:failure}), so objects move at the wrong time or not at all. STEP plans each event explicitly and verifies that it occurs, which mainly corrects when and where motion happens. Accordingly, STIoU improves in all six settings, by up to 0.092 on Wan2.2-VACE in I2V (0.1482 to 0.2403). SIoU, which ignores timing, changes less and even drops slightly on Wan2.2-VACE in V2V. The average gain is larger in I2V than in V2V (5.75 vs.\ 4.28 points). A single image leaves the upcoming event ambiguous, whereas a video prefix already shows the motion in progress, so an explicit event plan helps more in I2V. WMReward steers sampling with a reward on whole candidates. When the base generator rarely produces the expected event, the reward has little to steer toward, which likely explains its small gain on MiniMax H3 in I2V (36.42 to 36.63). Its reward also favors visual plausibility: on Wan2.2-VACE, it raises SIoU and WSIoU more than STEP does, but its STIoU stays below that of STEP. On MAGI-1 and MiniMax H3, WMReward attains a higher STIoU or a lower MSE in some settings, yet STEP achieves the highest score for every base generator in both settings.

\noindent\textbf{Qualitative results.}
Figure~\ref{fig:qualitative} shows representative cases. In the first I2V case, the grippers release a ball and a block onto a cushion. MiniMax H3 does not drop them at first. When they finally fall, both pass through the cushion. STEP drops both objects, which come to rest on the cushion. In the second I2V case, the statue rotates clockwise. MiniMax H3 rotates it counterclockwise, and its shadow does not match the statue. STEP keeps the clockwise rotation and a consistent shadow. In the first V2V case, a gripper releases one ball of a Newton's cradle. MiniMax H3 keeps the gripper closed at first. After the release, the collision does not pass the motion along the cradle. STEP releases the ball and passes the motion through the cradle as in the ground truth. In the second V2V case, a potato sinks into the liquid. MiniMax H3 makes it float up and vanish, and a second potato appears above the glass. STEP sinks the potato into the liquid as in the ground truth.

\subsubsection{Text-to-Video Generation}
\label{sec:exp_text}

\noindent\textbf{Quantitative results.}
Table~\ref{tab:main_phygen} reports results on PhyGenBench. Without a visual input, the base generators must infer the whole process from text, and they fail in different ways. Among the three base generators, Wan2.2-VACE has the lowest naturalness (0.4729), producing implausible processes such as an egg that keeps flying and never reaches the rock (Fig.~\ref{fig:qualitative}). STEP raises its naturalness to 0.7152. MiniMax H3 already shows most key phenomena (0.6833) but scores lower in physics order (0.4917). STEP raises its order score to 0.6146, consistent with planning events in order. WMReward brings a smaller gain on MiniMax H3 and even lowers the PCA on Wan2.2-VACE (0.4958 to 0.4667), where its order score also drops (0.4188 to 0.3938), since a reward on whole candidates does not specify which event should come next. On MAGI-1, the gains are smaller, and the thermal score drops slightly (0.4222 to 0.4000).

\noindent\textbf{Qualitative results.}
In the egg case of Fig.~\ref{fig:qualitative}, the egg should break on impact with the rock. MiniMax H3 makes the egg hit the rock and bounce off intact. In Wan2.2-VACE, the egg keeps flying through the air for the whole video and never reaches the rock. STEP produces the impact, and the egg breaks. In the paint case, the two colors should blend into a uniform mixture. MiniMax H3 drifts to blue, and Wan2.2-VACE leaves unmixed streaks. STEP blends the two colors into a uniform mixture.

\subsection{Ablation Studies and Analysis}
\label{sec:exp_ablations}

\begin{figure}[t]
  \centering
  \includegraphics[width=\linewidth]{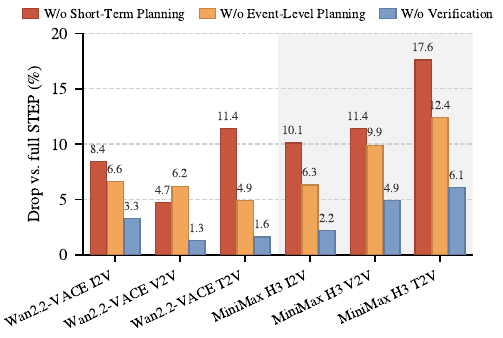}
  \caption{Ablation results on Wan2.2-VACE and MiniMax H3. Bars show the relative drop of each variant from the full STEP, measured by the Physics-IQ Verified score for I2V and V2V and by PCA for T2V.}
  \label{fig:ablation_bar}
\end{figure}

\begin{figure}[t]
  \centering
  \includegraphics[width=\linewidth]{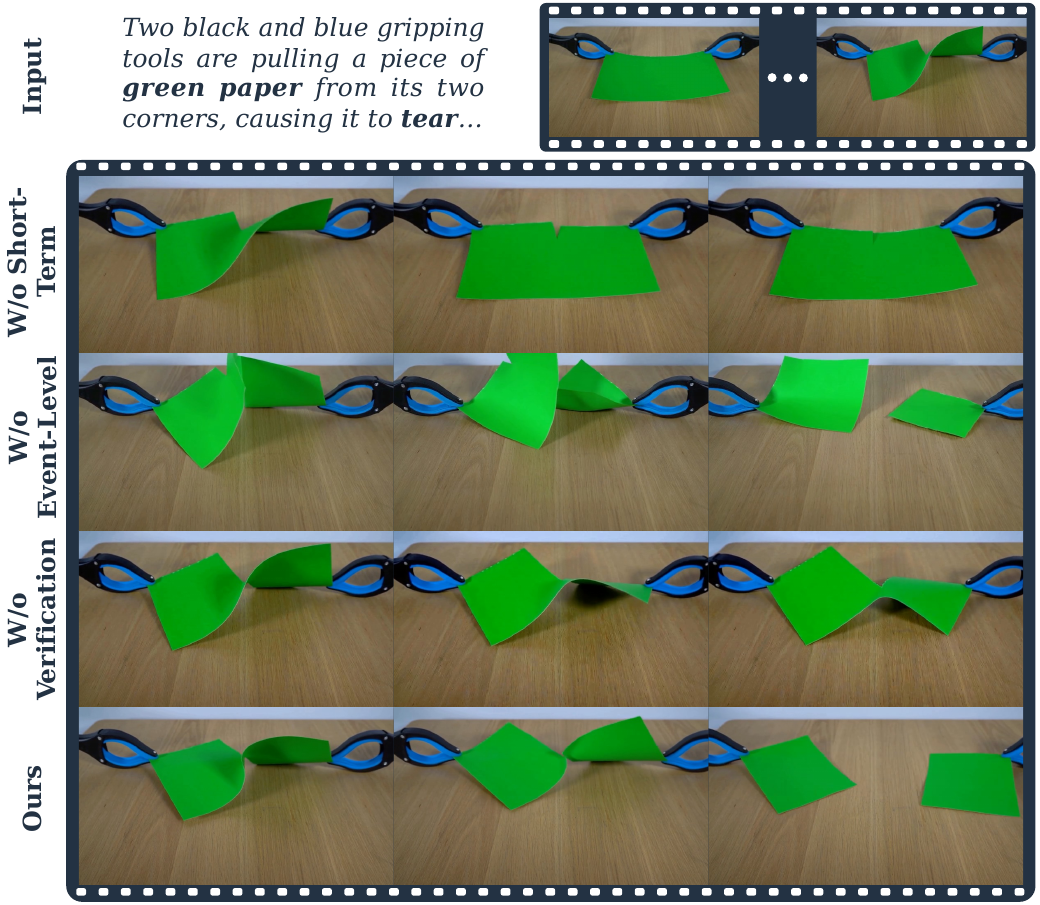}
  \caption{Qualitative ablation. Rows show w/o short-term planning, w/o event-level planning, w/o verification, and ours (full STEP), all on MiniMax H3 from the same input.}
  \label{fig:ablation}
\end{figure}

All three components of STEP contribute to its performance: removing any one of them lowers the score in every setting. We test this with three variants, each of which removes one component and keeps all other settings fixed. Figure~\ref{fig:ablation_bar} reports their results on Wan2.2-VACE and MiniMax H3, and Fig.~\ref{fig:ablation} visualizes the results of each variant.

\suppressfloats[t]
\subsubsection{Short-Term Planning}
\label{sec:exp_abl_horizon}

The w/o short-term planning variant calls the planner once at initialization and fixes per-second targets as absolute end positions, so later measurements cannot revise the plan. Both settings keep the same clip-wise generation, so the comparison isolates the planning horizon. This variant causes the largest drop in five of the six settings, ranging from 4.7\% to 17.6\%. In Fig.~\ref{fig:ablation}, the grippers only stretch and deform the paper, and the tear never occurs. Long-range predictions are unreliable: by the time the paper should tear, the fixed plan has drifted from the actual scene state and cannot be corrected, so the whole event is missing.

\subsubsection{Event-Level Planning}
\label{sec:exp_abl_event}

The w/o event-level planning variant removes the event list and the event history from the planner input, so the planner directly outputs per-object operations and coarse targets. Clip construction, latent guidance, and operation-level checks are kept. The drop ranges from 4.9\% to 12.4\%, and on Wan2.2-VACE in V2V it exceeds that of w/o short-term planning. In Fig.~\ref{fig:ablation}, the tear still occurs, showing that short-term motion control works on its own. However, the tear follows a wrong path, and the two pieces differ in size and shape. Without explicit events and event history, the planner can move objects but cannot ensure that the motion produces the state change the event requires.

\subsubsection{Verification}
\label{sec:exp_abl_verify}

The w/o verification variant generates a single candidate per round and accepts it without checking, but still re-measures the generated frames, so the loop remains closed. Its drop is the smallest, ranging from 1.3\% to 6.1\%, but it appears in every setting. In Fig.~\ref{fig:ablation}, the tear starts at the middle and partly spreads, but the paper is never fully torn. When a clip is generated poorly, for example when the tear does not spread far enough, the wrong state passes directly into the next round of planning. With all components, STEP completes the event: the tear spreads in the expected direction, and the paper splits into two pieces of similar size.

\section{Conclusion}
\label{sec:conclusion}

Video generators often fail at physical events: an expected event does not occur or is replaced by a different one. Methods that fix control for the whole clip before generation rely on long-range predictions, which become unreliable in later frames. We present STEP, a training-free closed-loop framework that plans only the next event for a short clip from the current scene state, converts it into a local control signal, and verifies the generated clip before continuing. STEP improves the physical plausibility of different frozen video generators across generation settings, and the ablations show that short-term planning, event-level planning, and verification each contribute to this improvement.

STEP depends on accurate object measurements and on how well the frozen generator follows local control. Its clip length is also fixed across scenes, although processes of different speeds may call for different lengths. Choosing the clip length adaptively is a direction for future work.

{
    \small
    \bibliographystyle{ieeenat_fullname}
    \bibliography{main}

\begin{thebibliography}{50}
\providecommand{\natexlab}[1]{#1}
\providecommand{\url}[1]{\texttt{#1}}
\expandafter\ifx\csname urlstyle\endcsname\relax
  \providecommand{\doi}[1]{doi: #1}\else
  \providecommand{\doi}{doi: \begingroup \urlstyle{rm}\Url}\fi

\bibitem[Bansal et~al.(2026)Bansal, Peng, Bitton, Goldenberg, Grover, and
  Chang]{bansal2026videophy2}
Hritik Bansal, Clark Peng, Yonatan Bitton, Roman Goldenberg, Aditya Grover, and
  Kai-Wei Chang.
\newblock {VideoPhy-2}: {A} challenging action-centric physical commonsense
  evaluation in video generation.
\newblock In \emph{International Conference on Learning Representations
  (ICLR)}, 2026.

\bibitem[Carion et~al.(2026)Carion, Gustafson, Hu, et~al.]{carion2026sam3}
Nicolas Carion, Laura Gustafson, Yuan-Ting Hu, et~al.
\newblock {SAM} 3: Segment anything with concepts.
\newblock In \emph{International Conference on Learning Representations
  (ICLR)}, 2026.

\bibitem[Chao et~al.(2026)Chao, Preechakul, Liu, and Bai]{chao2026seriality}
Jorge~Diaz Chao, Konpat Preechakul, Yuxi Liu, and Yutong Bai.
\newblock The seriality gap in video diffusion models.
\newblock arXiv preprint arXiv:2607.13031, 2026.

\bibitem[Chen et~al.(2026)Chen, Lin, and Zhang]{chen2026codeworldmodel}
Yiwen Chen, Guosheng Lin, and Chi Zhang.
\newblock {Code World Model}: Coding agent as world brain.
\newblock arXiv preprint arXiv:2608.25927, 2026.

\bibitem[Deng et~al.(2026)Deng, Pan, Zhang, et~al.]{deng2026rethinking}
Yufan Deng, Zilin Pan, Hongyu Zhang, et~al.
\newblock Rethinking video generation model for the embodied world.
\newblock In \emph{International Conference on Machine Learning (ICML)}, 2026.

\bibitem[Esmati et~al.(2026)Esmati, Nath, Hofmann, Nowrouzezahrai, Kahou, and
  Mirmehdi]{esmati2026invisible}
Parsa Esmati, Somjit Nath, Katja Hofmann, Derek Nowrouzezahrai, Samira~Ebrahimi
  Kahou, and Majid Mirmehdi.
\newblock The invisible hand of physics: When video diffusion models know more
  than they show.
\newblock arXiv preprint arXiv:2606.05328, 2026.

\bibitem[Feng et~al.(2026)Feng, Wang, Xu, Qian, Wang, Hou, Liu, Sun, Liu, and
  Wang]{feng2026newton}
Yuxiang Feng, Juncheng Wang, Chao Xu, Yijie Qian, Huihan Wang, Wenlong Hou,
  Yang Liu, Baigui Sun, Yong Liu, and Shujun Wang.
\newblock {NEWTON}: Agentic planning for physically grounded video generation.
\newblock arXiv preprint arXiv:2605.18396, 2026.

\bibitem[Foo et~al.(2026)Foo, Huang, Lattas, Moschoglou, Beeler, and
  Theobalt]{foo2026psivg}
Lin~Geng Foo, Mark~He Huang, Alexandros Lattas, Stylianos Moschoglou, Thabo
  Beeler, and Christian Theobalt.
\newblock Physical simulator in-the-loop video generation.
\newblock In \emph{IEEE/CVF Conference on Computer Vision and Pattern
  Recognition (CVPR)}, pages 4301--4311, 2026.

\bibitem[Guan et~al.(2026)Guan, Yang, Bao, Wang, Gu, Chen, Chen, Huo, and
  Gao]{guan2026agentic}
Xikang Guan, Shaocong Yang, Yanqi Bao, Can Wang, Zheng Gu, Zhengqing Chen, Kai
  Chen, Jing Huo, and Yang Gao.
\newblock Agentic physics-grounded video generation via memory-augmented
  in-context reasoning.
\newblock arXiv preprint arXiv:2610.23677, 2026.

\bibitem[HaCohen et~al.(2025)HaCohen, Chiprut, Brazowski,
  et~al.]{hacohen2025ltxvideo}
Yoav HaCohen, Nisan Chiprut, Benny Brazowski, et~al.
\newblock {LTX-Video}: Realtime video latent diffusion.
\newblock arXiv preprint arXiv:2501.00103, 2025.

\bibitem[HaCohen et~al.(2026)HaCohen, Brazowski, Chiprut,
  et~al.]{hacohen2026ltx2}
Yoav HaCohen, Benny Brazowski, Nisan Chiprut, et~al.
\newblock {LTX-2}: Efficient joint audio-visual foundation model.
\newblock arXiv preprint arXiv:2601.03233, 2026.

\bibitem[Han et~al.(2026)Han, Kang, Jun, Chen, Yang, and
  Hwang]{han2026physics2steps}
Woojung Han, Seil Kang, Youngjun Jun, Min-Hung Chen, Fu-En Yang, and Seong~Jae
  Hwang.
\newblock Physics in 2-steps: Locking motion priors before visual refinement
  erases them.
\newblock arXiv preprint arXiv:2606.06361, 2026.

\bibitem[Hassan et~al.(2026)Hassan, Messaoud, Li, and Alahi]{hassan2026proprio}
Mariam Hassan, Kaouther Messaoud, Wuyang Li, and Alexandre Alahi.
\newblock {Proprio}: Latent self-scoring and inference-time refinement for
  physically plausible video generation.
\newblock arXiv preprint arXiv:2605.28230, 2026.

\bibitem[Huang et~al.(2025)Huang, Wang, Lin, Kim, Omidshafiei, Yoon, Zhang, and
  Bansal]{huang2025sketchverify}
Yidong Huang, Zun Wang, Han Lin, Dong-Ki Kim, Shayegan Omidshafiei, Jaehong
  Yoon, Yue Zhang, and Mohit Bansal.
\newblock Planning with sketch-guided verification for physics-aware video
  generation.
\newblock arXiv preprint arXiv:2511.17450, 2025.

\bibitem[Jiang et~al.(2025)Jiang, Han, Mao, Zhang, Pan, and Liu]{jiang2025vace}
Zeyinzi Jiang, Zhen Han, Chaojie Mao, Jingfeng Zhang, Yulin Pan, and Yu Liu.
\newblock {VACE}: All-in-one video creation and editing.
\newblock In \emph{IEEE/CVF International Conference on Computer Vision
  (ICCV)}, pages 17191--17202, 2025.

\bibitem[Kang et~al.(2025)Kang, Yue, Lu, Lin, Zhao, Wang, Huang, and
  Feng]{kang2025videoworldmodel}
Bingyi Kang, Yang Yue, Rui Lu, Zhijie Lin, Yang Zhao, Kaixin Wang, Gao Huang,
  and Jiashi Feng.
\newblock How far is video generation from world model: {A} physical law
  perspective.
\newblock In \emph{International Conference on Machine Learning (ICML)}, 2025.

\bibitem[Kazimi et~al.(2026)Kazimi, Borse, Hayat, Porikli, and
  Yanardag]{kazimi2026hiphy}
Tahira Kazimi, Shubhankar Borse, Munawar Hayat, Fatih Porikli, and Pinar
  Yanardag.
\newblock {HiPhy}: Hierarchical alignment for physically-plausible
  multi-principle video generation.
\newblock arXiv preprint arXiv:2610.02197, 2026.

\bibitem[Ki et~al.(2026)Ki, Zhang, Zheng, Wang, Zhu, Chen, Zhou, and
  Lu]{ki2026stateworld}
Chan~Yan Ki, Yanran Zhang, Wenzhao Zheng, Xiaofeng Wang, Zheng Zhu, Lei Chen,
  Jie Zhou, and Jiwen Lu.
\newblock State world models: Execute the rules, generate the pixels.
\newblock arXiv preprint arXiv:2610.26811, 2026.

\bibitem[Kizil et~al.(2026)Kizil, Sanli, Mitra, Erdem, Erdem, and
  Ceylan]{kizil2026lamp}
Muhammed~Burak Kizil, Enes Sanli, Niloy~J. Mitra, Erkut Erdem, Aykut Erdem, and
  Duygu Ceylan.
\newblock {LAMP}: Language-assisted motion planning for controllable video
  generation.
\newblock In \emph{IEEE/CVF Conference on Computer Vision and Pattern
  Recognition (CVPR)}, pages 12829--12838, 2026.

\bibitem[Le et~al.(2025)Le, Zhu, Kalogeiton, and Samaras]{le2025gravity}
Minh-Quan Le, Yuanzhi Zhu, Vicky Kalogeiton, and Dimitris Samaras.
\newblock What about gravity in video generation? post-training {Newton}'s laws
  with verifiable rewards.
\newblock arXiv preprint arXiv:2512.00425, 2025.

\bibitem[Li et~al.(2026)Li, Hao, Li, Yi, Rosin, and Lai]{li2026physagent}
Qirui Li, Jinkun Hao, Yibo Li, Ran Yi, Paul~L. Rosin, and Yu-Kun Lai.
\newblock {PhysAgent}: Reflective agentic physics control for physically
  plausible video generation.
\newblock arXiv preprint arXiv:2607.16355, 2026.

\bibitem[Lian et~al.(2024)Lian, Shi, Yala, Darrell, and
  Li]{lian2024llmgrounded}
Long Lian, Baifeng Shi, Adam Yala, Trevor Darrell, and Boyi Li.
\newblock {LLM}-grounded video diffusion models.
\newblock In \emph{International Conference on Learning Representations
  (ICLR)}, 2024.

\bibitem[Liu et~al.(2024)Liu, Ren, Gupta, and Wang]{liu2024physgen}
Shaowei Liu, Zhongzheng Ren, Saurabh Gupta, and Shenlong Wang.
\newblock {PhysGen}: Rigid-body physics-grounded image-to-video generation.
\newblock In \emph{European Conference on Computer Vision (ECCV)}, pages
  360--378, 2024.

\bibitem[Liu et~al.(2026)Liu, Chen, Li, Wang, Yu, and Wu]{liu2026realwonder}
Wei Liu, Ziyu Chen, Zizhang Li, Yue Wang, Hong-Xing Yu, and Jiajun Wu.
\newblock {RealWonder}: Real-time physical action-conditioned video generation.
\newblock arXiv preprint arXiv:2603.05449, 2026.

\bibitem[Liu et~al.(2025)Liu, Zhao, Wen, Dai, and Huang]{liu2025bootstrapping}
Yang Liu, Xilin Zhao, Peisong Wen, Siran Dai, and Qingming Huang.
\newblock Bootstrapping physics-grounded video generation through {VLM}-guided
  iterative self-refinement.
\newblock arXiv preprint arXiv:2511.20280, 2025.

\bibitem[Meng et~al.(2025)Meng, Liao, Tan, Lu, Shao, Zhang, Cheng, Li, and
  Luo]{meng2025phygenbench}
Fanqing Meng, Jiaqi Liao, Xinyu Tan, Quanfeng Lu, Wenqi Shao, Kaipeng Zhang, Yu
  Cheng, Dianqi Li, and Ping Luo.
\newblock Towards world simulator: Crafting physical commonsense-based
  benchmark for video generation.
\newblock In \emph{International Conference on Machine Learning (ICML)}, 2025.

\bibitem[Meng et~al.(2026)Meng, Luo, and Liu]{meng2026phymagic}
Siwei Meng, Yawei Luo, and Ping Liu.
\newblock {PhyMAGIC}: Physical motion-aware generative inference with
  confidence-guided {VLM}.
\newblock In \emph{European Conference on Computer Vision (ECCV)}, pages
  57--73, 2026.

\bibitem[{MiniMax}(2026)]{minimax2026h3}
{MiniMax}.
\newblock {MiniMax H3}: An open model breaking the boundaries between tasks and
  modalities.
\newblock \url{https://www.minimax.io/blog/minimax-h3}, 2026.

\bibitem[Motamed et~al.(2026)Motamed, Culp, Swersky, Jaini, and
  Geirhos]{motamed2026physicsiq}
Saman Motamed, Laura Culp, Kevin Swersky, Priyank Jaini, and Robert Geirhos.
\newblock Do generative video models understand physical principles?
\newblock In \emph{IEEE/CVF Winter Conference on Applications of Computer
  Vision (WACV)}, pages 948--958, 2026.

\bibitem[Narayanan et~al.(2026)Narayanan, Jiang, Narasimhan, and
  Chandraker]{narayanan2026phyco}
Sriram Narayanan, Ziyu Jiang, Srinivasa~G. Narasimhan, and Manmohan Chandraker.
\newblock {PhyCo}: Learning controllable physical priors for generative motion.
\newblock In \emph{IEEE/CVF Conference on Computer Vision and Pattern
  Recognition (CVPR)}, pages 41892--41902, 2026.

\bibitem[Nguyen et~al.(2026)Nguyen, Le, Le, Nguyen, Tran, and
  Do]{nguyen2026physplan}
Minh-Loi Nguyen, Xuan-Vu Le, Trung-Nghia Le, Tam~V. Nguyen, Minh-Triet Tran,
  and Thanh-Toan Do.
\newblock {PhysPlan}: Grounded physical state reasoning and graph-guided
  optimization for physically plausible video generation.
\newblock arXiv preprint arXiv:2609.13006, 2026.

\bibitem[{NVIDIA} et~al.(2025){NVIDIA}, Agarwal, Ali, Bala,
  et~al.]{agarwal2025cosmos}
{NVIDIA}, Niket Agarwal, Arslan Ali, Maciej Bala, et~al.
\newblock {Cosmos} world foundation model platform for physical {AI}.
\newblock arXiv preprint arXiv:2501.03575, 2025.

\bibitem[{NVIDIA} et~al.(2026){NVIDIA}, Aditi, Agarwal, Ali,
  et~al.]{aditi2026cosmos3}
{NVIDIA}, Aditi, Niket Agarwal, Arslan Ali, et~al.
\newblock {Cosmos} 3: Omnimodal world models for physical {AI}.
\newblock arXiv preprint arXiv:2606.02800, 2026.

\bibitem[R{\"a}dsch et~al.(2026)R{\"a}dsch, Asano, Kuehne, Bauer, Jaini,
  Geirhos, and L{\"u}th]{radsch2026physicsiqverified}
Tim R{\"a}dsch, Yuki~M. Asano, Hilde Kuehne, Stefan Bauer, Priyank Jaini,
  Robert Geirhos, and Carsten~T. L{\"u}th.
\newblock Physics-{IQ} {Verified}.
\newblock arXiv preprint arXiv:2606.18943, 2026.

\bibitem[{Sand.ai} et~al.(2025){Sand.ai}, Teng, Jia, Sun,
  et~al.]{teng2025magi1}
{Sand.ai}, Hansi Teng, Hongyu Jia, Lei Sun, et~al.
\newblock {MAGI-1}: Autoregressive video generation at scale.
\newblock arXiv preprint arXiv:2505.13211, 2025.

\bibitem[Shen et~al.(2026)Shen, Xiong, Yu, and Lourentzou]{shen2026phantom}
Ying Shen, Jerry Xiong, Tianjiao Yu, and Ismini Lourentzou.
\newblock {PHANTOM}: Physics-infused video generation via joint modeling of
  visual and latent physical dynamics.
\newblock In \emph{IEEE/CVF Conference on Computer Vision and Pattern
  Recognition (CVPR)}, pages 11185--11194, 2026.

\bibitem[Tan et~al.(2026)Tan, Jiang, Li, Zong, Xie, Yang, and
  Jiang]{tan2026physmotion}
Xiyang Tan, Ying Jiang, Xuan Li, Zeshun Zong, Tianyi Xie, Yin Yang, and
  Chenfanfu Jiang.
\newblock {PhysMotion}: Physics-grounded dynamics from a single image.
\newblock In \emph{International Conference on 3D Vision (3DV)}, pages
  806--818, 2026.

\bibitem[Tang et~al.(2026)Tang, Zhong, and Ding]{tang2026seeking}
Chujun Tang, Lei Zhong, and Fangqiang Ding.
\newblock Seeking physics in diffusion noise.
\newblock arXiv preprint arXiv:2603.14294, 2026.

\bibitem[{Team Wan} et~al.(2025){Team Wan}, Wang, Ai, Wen, et~al.]{wang2025wan}
{Team Wan}, Ang Wang, Baole Ai, Bin Wen, et~al.
\newblock {Wan}: Open and advanced large-scale video generative models.
\newblock arXiv preprint arXiv:2503.20314, 2025.

\bibitem[{Tencent Hunyuan Foundation Model
  Team}(2025)]{tencent2025hunyuanvideo15}
{Tencent Hunyuan Foundation Model Team}.
\newblock {HunyuanVideo} 1.5 technical report.
\newblock arXiv preprint arXiv:2511.18870, 2025.

\bibitem[Wang et~al.(2025)Wang, Chen, Huang, Dou, Liu, Gu, and
  Liu]{wang2025physctrl}
Chen Wang, Chuhao Chen, Yiming Huang, Zhiyang Dou, Yuan Liu, Jiatao Gu, and
  Lingjie Liu.
\newblock {PhysCtrl}: Generative physics for controllable and physics-grounded
  video generation.
\newblock In \emph{Advances in Neural Information Processing Systems
  (NeurIPS)}, 2025.

\bibitem[Wang et~al.(2026{\natexlab{a}})Wang, Cai, Chen,
  et~al.]{wang2026codeasworlds}
Hanyang Wang, Yimo Cai, Weiliang Chen, et~al.
\newblock Code as worlds: Agentic discovery of executable world representations
  for physical reasoning.
\newblock arXiv preprint arXiv:2608.27549, 2026{\natexlab{a}}.

\bibitem[Wang et~al.(2026{\natexlab{b}})Wang, Hu, Wang, Chen, Liu, Li, and
  Lei]{wang2026eventcausal}
Zixuan Wang, Yixin Hu, Haolan Wang, Feng Chen, Yan Liu, Wen Li, and Yinjie Lei.
\newblock Chain of event-centric causal thought for physically plausible video
  generation.
\newblock In \emph{IEEE/CVF Conference on Computer Vision and Pattern
  Recognition (CVPR)}, pages 38122--38131, 2026{\natexlab{b}}.

\bibitem[Xue et~al.(2025)Xue, Yin, Yang, and Gao]{xue2025phyt2v}
Qiyao Xue, Xiangyu Yin, Boyuan Yang, and Wei Gao.
\newblock {PhyT2V}: {LLM}-guided iterative self-refinement for physics-grounded
  text-to-video generation.
\newblock In \emph{IEEE/CVF Conference on Computer Vision and Pattern
  Recognition (CVPR)}, pages 18826--18836, 2025.

\bibitem[Yang et~al.(2025{\natexlab{a}})Yang, Li, Zhang, et~al.]{yang2025vlipp}
Xindi Yang, Baolu Li, Yiming Zhang, et~al.
\newblock {VLIPP}: Towards physically plausible video generation with vision
  and language informed physical prior.
\newblock In \emph{IEEE/CVF International Conference on Computer Vision
  (ICCV)}, pages 12360--12370, 2025{\natexlab{a}}.

\bibitem[Yang et~al.(2025{\natexlab{b}})Yang, Teng, Zheng,
  et~al.]{yang2025cogvideox}
Zhuoyi Yang, Jiayan Teng, Wendi Zheng, et~al.
\newblock {CogVideoX}: Text-to-video diffusion models with an expert
  transformer.
\newblock In \emph{International Conference on Learning Representations
  (ICLR)}, 2025{\natexlab{b}}.

\bibitem[Ye et~al.(2026)Ye, Ge, Zheng, et~al.]{ye2026worldaction}
Seonghyeon Ye, Yunhao Ge, Kaiyuan Zheng, et~al.
\newblock World action models are zero-shot policies.
\newblock arXiv preprint arXiv:2602.15922, 2026.

\bibitem[Yuan et~al.(2026{\natexlab{a}})Yuan, Zhang, Friedrich, Beltran-Velez,
  Hall, Hemmat, Han, Ballas, Drozdzal, and Romero-Soriano]{yuan2026wmreward}
Jianhao Yuan, Xiaofeng Zhang, Felix Friedrich, Nicolas Beltran-Velez, Melissa
  Hall, Reyhane~Askari Hemmat, Xiaochuang Han, Nicolas Ballas, Michal Drozdzal,
  and Adriana Romero-Soriano.
\newblock Inference-time physics alignment of video generative models with
  latent world models.
\newblock In \emph{IEEE/CVF Conference on Computer Vision and Pattern
  Recognition (CVPR)}, pages 16118--16129, 2026{\natexlab{a}}.

\bibitem[Yuan et~al.(2026{\natexlab{b}})Yuan, Wang, Wickremasinghe, Nadir, Ma,
  and Chan]{yuan2026newtongen}
Yu Yuan, Xijun Wang, Tharindu Wickremasinghe, Zeeshan Nadir, Bole Ma, and
  Stanley Chan.
\newblock {NewtonGen}: Physics-consistent and controllable text-to-video
  generation via neural {Newtonian} dynamics.
\newblock In \emph{International Conference on Learning Representations
  (ICLR)}, 2026{\natexlab{b}}.

\bibitem[Zhao et~al.(2026)Zhao, Li, He, and Wu]{zhao2026phyrpr}
Yibo Zhao, Hengjia Li, Xiaofei He, and Boxi Wu.
\newblock {PhyRPR}: Training-free physics-constrained video generation.
\newblock arXiv preprint arXiv:2601.09255, 2026.

\end{thebibliography}
}

\clearpage
\appendix
\section*{\centering Appendix}
This appendix documents the inference settings for the video generators and comparison methods in the main text and explains how STEP plans, generates, and verifies successive clips across base generators. We then analyze the inference time and external model API usage of STEP, and discuss its limitations through failure cases and directions for future work. Additional image-to-video, video-to-video, and text-to-video examples, with side-by-side comparisons of MiniMax H3 and STEP applied to MiniMax H3, are available on the project page linked on the first page.

\setlength{\stripsep}{6pt plus 2pt minus 2pt}
\begin{strip}
\centering
\captionsetup{hypcap=false}
\captionof{table}{Sampling settings for the video generators in the main text.}
\label{tab:supp-sampling}
\small
\setlength{\tabcolsep}{3pt}
\renewcommand{\arraystretch}{1.1}
\begin{tabular*}{\textwidth}{@{\extracolsep{\fill}}l c c c c c >{\raggedright\arraybackslash}p{0.30\textwidth}@{}}
\toprule
 & & & \multicolumn{2}{c}{Resolution} & & \\
\cmidrule(lr){4-5}
Model & Frames & FPS & I2V/T2V & V2V & Steps & Guidance \\
\midrule
HunyuanVideo 1.5 & 121 & 24 & $848{\times}480$ & -- & 50 & 6.0 \\
CogVideoX-5B / 5B-I2V & 41 & 8 & $720{\times}480$ & -- & 50 & 6.0 \\
LTX-2.5 22B & 121 & 24 & $768{\times}512$ & $896{\times}512$ & 40 & 3.0 \\
LTX-Video 0.9.8 13B & 81 & 16 & $736{\times}480$ & $864{\times}480$ & $27+13$ & Stage 1: $[1,1,6,8,6,1,1]$\newline
  Stage 2: $1.0$ \\
NewtonGen (T2V only) & 41 & 8 & $720{\times}480$ & -- & 30 & 6.0 \\
Wan2.2-VACE-Fun-A14B & 81 & 16 & $720{\times}480$ & $720{\times}480$ & 30 & 5.0 \\
MAGI-1 4.5B & 120 & 24 & $848{\times}480$ & $848{\times}480$ & 64 &
  Text $[7.5,7.5,7.5,0,0]$\newline
  Previous chunk $[1.5,1.5,1.5,1.0,1.0]$ \\
MiniMax H3 & 124 & 24 & $768{\times}512$ & $768{\times}512$ & 50 & \textemdash \\
\bottomrule
\end{tabular*}

\smallskip
{\raggedright\footnotesize
CogVideoX uses task-specific checkpoints.
LTX-2.5 and LTX-Video 0.9.8 use the dev checkpoints. The step counts exclude the skipped updates in each denoising stage.
MAGI-1 uses sampling time points $[0,0.0217,0.1,0.3,0.999]$ for both guidance schedules.
MiniMax H3 incorporates guidance into the model weights and exposes no separate guidance scale.
\par}
\end{strip}

\begin{table*}[t]
\centering
\caption{Additional inference settings for WMReward and STEP.}
\label{tab:supp-step}
\small
\begin{tabularx}{\textwidth}{@{}l >{\raggedright\arraybackslash}X@{}}
\toprule
Method & Inference settings \\
\midrule
Wan2.2-VACE + WMReward & 16 candidates, physics reward guidance scale 0.001. \\
MAGI-1 + WMReward & 16 candidates, physics reward guidance scale 0.001. \\
MiniMax H3 + WMReward & 16 candidates, physics reward guidance scale 0.001. \\
\midrule
Wan2.2-VACE + STEP & Five rounds with 16 retained frames per round. I2V/T2V condition on 1, 17, 33, 33, and 33 frames across rounds. V2V conditions on 32 frames in each round. Anchoring applies during the first $k_0=14$ denoising updates. \\
MAGI-1 + STEP & Five rounds with 24 retained frames per round. I2V/T2V condition on 1, 24, 48, 48, and 48 frames across rounds. V2V conditions on 48 frames in each round. Anchoring applies while $\sigma_{k+1}\geq0.93$, which sets $k_0$. \\
MiniMax H3 + STEP & Five rounds retaining 24, 24, 24, 24, and 28 frames, where the last round retains 28 frames for chunk alignment. I2V/T2V condition on 1, 22, 39, 39, and 39 frames across rounds. V2V conditions on 39 frames in each round. Anchoring applies while $\sigma_{k+1}\geq0.93$, which sets $k_0$. \\
\bottomrule
\end{tabularx}
\end{table*}

\section{Implementation Details}

\subsection{Base Generators and Comparison Methods}

Table~\ref{tab:supp-sampling} lists the video generators evaluated on Physics-IQ Verified and PhyGenBench. We denote image-to-video, video-to-video, and text-to-video by I2V, V2V, and T2V, respectively. STEP keeps the weights of Wan2.2-VACE, MAGI-1, and MiniMax H3 frozen and uses the sampling settings of the corresponding base generator. Frames denotes the number of output frames retained after decoding. Outputs are trimmed to five seconds.

\subsection{WMReward and STEP Inference}

\noindent\textbf{WMReward.}
Across the three base generators, WMReward uses gradients of a world model's physics reward to guide denoising. It generates 16 candidates per input and selects the video with the highest reward. On Wan2.2-VACE, the V-JEPA 2 prediction loss is evaluated on the decoded clean video estimate, and its gradient with respect to the noisy video latents corrects the denoising velocity. On MAGI-1, it is applied chunk by chunk at selected autoregressive denoising steps: the V-JEPA 2 prediction loss is backpropagated through the decoded clean estimate to the current noisy chunk latent to correct its denoising velocity. On MiniMax H3, the same loss is evaluated over temporal windows of the decoded clean video estimate, and its gradient with respect to the noisy video latents corrects the velocity before each scheduler update.

\noindent\textbf{STEP.}
In each round, STEP anchors early denoising to coarse motion specified by the event plan. After generating a clip, STEP checks whether the planned operations occurred and whether the constraints affected by these operations hold. If a check fails, STEP resamples with a different random seed while keeping the context and event plan fixed. Table~\ref{tab:supp-step} summarizes the additional inference settings.

\subsection{Conditioning and Reuse}

In each round, STEP constructs the control conditions and encodes the prompt once. Retries reuse both and draw only fresh initial noise.

In V2V, the generator conditions on recent frames and retains about one second of new frames per round (28 frames in the last MiniMax H3 round). The planner and verifier retain their observation windows. In I2V and T2V, the generator starts from a single image and uses more history in later rounds. Table~\ref{tab:supp-step} lists the conditioning frames supplied to the generator in each round, which follow the chunk constraints of each base generator.

\subsection{Verification and Candidate Selection}

We take Wan2.2-VACE + STEP in V2V as an example. In each round, the generator conditions on 32 frames and generates 17 new frames, and STEP verifies the first 16, which form the candidate clip. A passed check ends sampling for the round. A failed check triggers up to $N=2$ additional attempts with different seeds and the same context and plan. A check is unmeasurable when $\Phi$ cannot obtain the quantity it requires from the generated clip. Table~\ref{tab:supp-checks} lists the pass criteria of each check. An unmeasurable result is not counted as a pass, but it does not trigger resampling. If all attempts fail, STEP accepts the candidate closest to the plan, ranking candidates by whether all checks pass, whether the planned event is observed, and the number of passed checks minus the number of failed checks, in that order. Ties favor the earlier attempt. The accepted clip extends the video, and an event enters the event history only if its checks pass for all planned entities. Table~\ref{tab:supp-rollout} traces two rounds of one example.

\begin{table}[!htbp]
\centering
\caption{Pass criteria of the verification checks. Start values are measured on the last context frame, i.e., the scene state at the start of the round. Coordinates are normalized by image width and height. For Translate, the displacement is measured from the start mask centroid to the median centroid over the terminal window, the last $0.25$\,s of the clip.}
\label{tab:supp-checks}
\footnotesize
\setlength{\tabcolsep}{3pt}
\begin{tabularx}{\columnwidth}{@{}l >{\raggedright\arraybackslash}X@{}}
\toprule
Check & Pass criterion \\
\midrule
Translate & Target: the rendered control displacement, or the planned displacement if unavailable. Displacement $\geq0.03$ with cosine similarity $\geq0.5$ to the target direction. If the target displacement is $<0.03$, displacement $\leq0.05$. \\
Scale & $\log r/\log(f^2)\geq0.5$, where $f$ is the target linear scale and $r$ the measured mask-area ratio. Unmeasurable if $f\leq0$ or $|f-1|<0.1$. \\
Rotate & Accumulated rotation of the mask principal axis has the planned sign and reaches $50\%$ of the target angle. Unmeasurable if the target is below $10^\circ$ or any mask has an axis ratio below $1.2$. \\
Appear / disappear & A mask is visible if it covers at least $0.02\%$ of the frame. The entity appears if at least half of the frames in the last $0.25$\,s are visible, and disappears otherwise. \\
Color & Distance from the mean masked RGB color to the target color drops by at least $50\%$ from the start to the visible terminal frames. Unmeasurable if the initial distance is below $30$ (on a $0$--$255$ scale). \\
Support & Fails if, in the last visible frame, the bottom edge of the entity crosses the support limit by more than $0.03$ of the image height. The limit comes from the support surface, container rim, or container interior at the start of the round. \\
\bottomrule
\end{tabularx}
\end{table}

\begin{table}[!htbp]
\centering
\caption{Example rollout of Wan2.2-VACE + STEP in V2V on Physics-IQ Verified sample 0008, in which a ball rolls leftward. Each row shows the plan of one round, the verification of its candidates, and the accepted clip.}
\label{tab:supp-rollout}
\small
\begin{tabularx}{\columnwidth}{@{}c >{\raggedright\arraybackslash}X@{}}
\toprule
Round & Plan, verification, and accepted clip \\
\midrule
4 & Plan: the ball continues rolling left and leaves the view. Candidate 1 fails the translation check. Candidate 2, in which the ball leaves the frame, passes and is accepted. \\
5 & The updated scene state marks the ball as out of frame, so the plan adds no new operation. Candidate 1 passes the checks that the remaining objects stay still and is accepted. \\
\bottomrule
\end{tabularx}
\end{table}

\subsection{Evaluation Window and Alignment}

On Physics-IQ Verified, we trim all generated continuations to the first five seconds and exclude the conditioning frames from scoring. The benchmark provides test videos and reference masks at multiple frame rates, including 8, 16, and 24 FPS. For each submission, the evaluator uses references at the matching frame rate and resizes generated frames, reference frames, and masks to $960{\times}540$ for scoring. This protocol is fixed by the official evaluator and identical for all methods, so the metrics are computed on the same spatial grid and time axis regardless of each generator's native resolution and frame rate.

\subsection{Manual Annotation of Failure Types}

Figure~\ref{fig:failure}(b) reports the failure-type proportions of Wan2.2-VACE and MiniMax H3. To obtain them, we randomly sample 100 generated V2V videos per model from Wan2.2-VACE and MiniMax H3 on Physics-IQ Verified and compare each video with its ground-truth video. We first check whether the expected event occurs correctly. For example, we label a video as an event error if the ball bounces or moves in a circle when it should roll in a straight line. If the direction and type of motion match the ground truth but the object moves at a different speed or occupies a different position at the same time, we label it as a speed or position error. Because an event error can itself cause deviations in speed and position, a video exhibiting both types of error is counted only in the event category. Two annotators applied these criteria, each labeling the videos of one model. Percentages are computed over all 100 sampled videos per model.

\section{Overhead Analysis}
\subsection{Inference Time Overhead}

\begin{table}[!htbp]
\centering
\caption{Mean end-to-end generation time (s) per video on a single GPU (NVIDIA L20 for Wan2.2-VACE and MAGI-1, A100 for MiniMax H3). STEP includes all API requests, segmentation, and resampling. For WMReward, the serial time is estimated as the measured mean time per candidate $\times$ 16.}
\label{tab:supp-runtime}
\small
\setlength{\tabcolsep}{3pt}
\setlength{\aboverulesep}{1pt}
\setlength{\belowrulesep}{1pt}
\begin{tabular*}{\columnwidth}{@{\extracolsep{\fill}}lrrr@{}}
\toprule
Method & I2V & V2V & T2V \\
\midrule
Wan2.2-VACE & 564.6 & 1,009.3 & 559.7 \\
\quad + STEP & 1,082.7 & 1,257.3 & 1,004.2 \\
\quad + WMReward & 1,276.4 $\times$ 16 & 2,321.1 $\times$ 16 & 1,447.8 $\times$ 16 \\
\midrule
MAGI-1 4.5B & 2,161.2 & 1,913.7 & 1,760.7 \\
\quad + STEP & 2,659.6 & 2,715.8 & 2,266.8 \\
\quad + WMReward & 4,120.7 $\times$ 16 & 3,832.2 $\times$ 16 & 3,925.3 $\times$ 16 \\
\midrule
MiniMax H3 & 358.5 & 587.4 & 303.7 \\
\quad + STEP & 938.1 & 1,086.3 & 1,169.1 \\
\quad + WMReward & 883.5 $\times$ 16 & 1,376.3 $\times$ 16 & 877.1 $\times$ 16 \\
\bottomrule
\end{tabular*}
\end{table}

WMReward generates 16 full videos per input, each with reward-gradient backpropagation during denoising, and keeps the highest-scoring one. STEP instead keeps the accepted history and adds computation only for the current clip when it fails, concentrating the search budget on local errors. As Table~\ref{tab:supp-runtime} shows, it takes $1.23$--$1.92\times$ the base generator time on Wan and MAGI and $1.85$--$3.85\times$ on H3.

\subsection{External Model API Usage}
STEP averages $6.24$--$7.89$ requests per video (Table~\ref{tab:supp-api}). After the initial physical specification, each planning call adapts the targets to the generated scene state. These calls are necessary: in the main text, the w/o short-term planning variant plans once at initialization and keeps the same clip-wise generation, yet it causes the largest drop in five of the six settings. Retries reuse the plan, so they add sampling time without another planning call. Requests beyond one specification call and one planning call per round come from replanning and from re-sending requests whose responses fail to parse. The separate call to generate the T2V first frame is excluded.

\begin{table}[!htbp]
\centering
\caption{Mean external language or vision-language model API requests per output video.}
\label{tab:supp-api}
\small
\setlength{\tabcolsep}{3pt}
\setlength{\aboverulesep}{1pt}
\setlength{\belowrulesep}{1pt}
\begin{tabular*}{\columnwidth}{@{\extracolsep{\fill}}lrrr@{}}
\toprule
Method & I2V & V2V & T2V \\
\midrule
Wan2.2-VACE + STEP & 7.89 & 6.24 & 6.80 \\
MAGI-1 4.5B + STEP & 6.58 & 6.43 & 6.46 \\
MiniMax H3 + STEP & 6.35 & 6.40 & 6.44 \\
\bottomrule
\end{tabular*}
\end{table}

\begin{figure*}[t]
\centering
\includegraphics[width=\textwidth]{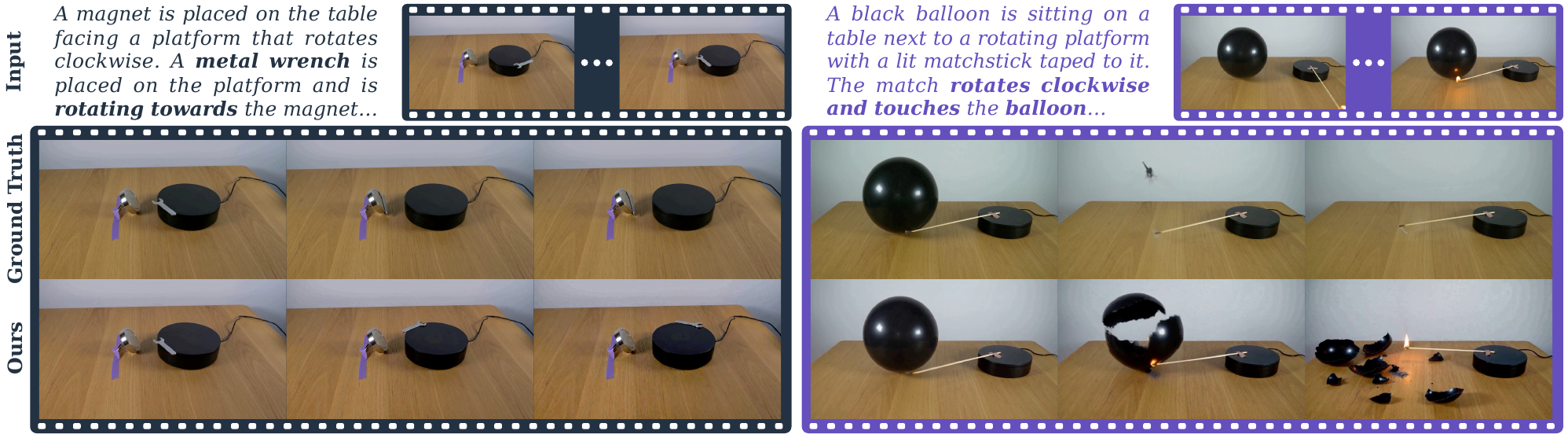}
\caption{Two failure cases of STEP. In the left example, the wrench remains on the turntable instead of moving toward the magnet. In the right example, the balloon bursts on contact with the lit match, but large fragments remain.}
\label{fig:supp-failures}
\end{figure*}

\section{Discussion}
\subsection{Failure Case Analysis}

\noindent\textbf{Magnet and wrench.}
In the ground-truth video, the turntable brings the wrench close to the magnet, and magnetic attraction pulls the wrench off the turntable. In the STEP result (Fig.~\ref{fig:supp-failures}, left), the wrench rotates with the turntable and remains on it. Either the VLM or the SAM 3 segmentation misidentified the magnet, so the plan omitted the transition from rotating with the turntable to moving toward the magnet. The omitted event had no corresponding motion control or verification check. As a result, the verifier could not flag the wrench's failure to leave the turntable.

\noindent\textbf{Balloon rupture.}
STEP generates the contact between the lit match and the balloon, followed by rupture (Fig.~\ref{fig:supp-failures}, right). The generated balloon leaves several large fragments, whereas the ground truth shows the balloon membrane rapidly retracting. This discrepancy likely reflects a limitation in how the base generator models the rupture of thin materials. The event plan can specify when the match touches the balloon and that rupture follows. The base generator determines where the membrane tears, how it retracts, and how large the fragments are. The current coarse motion controls provide little leverage over those details even when the sequence of events is correct.

\subsection{Future Work}
STEP consistently improves the physical plausibility of frozen video generators across image-to-video, video-to-video, and text-to-video generation, and opens up promising avenues for future exploration. Its closed loop rests on three components that can each be extended. The physical specification is built once at initialization, so an entity or event missed at this stage cannot be planned or verified later; revising the specification when the measured scene state contradicts it would let the loop recover such omissions. The scene state, in turn, is measured from masks on the image plane, where occlusion and motion along the viewing direction leave some quantities unmeasurable; tracking entities through occlusion and adding cues such as relative depth would make more events verifiable. Finally, every round uses the same clip length, although contact and abrupt state changes benefit from frequent feedback while steady motion does not; choosing the clip length adaptively would concentrate planning and verification where they matter most and further reduce inference cost.

\end{document}